\documentclass{article}
\usepackage[utf8]{inputenc}
\usepackage{newunicodechar}
\newunicodechar{♫}{\textmusicalnote}
\usepackage{iclr2027_conference,times}
\iclrfinalcopy 

\usepackage{amsmath,amsfonts,bm}

\def\eqref#1{equation~\ref{#1}}

\def\1{\bm{1}}

\DeclareMathAlphabet{\mathsfit}{\encodingdefault}{\sfdefault}{m}{sl}
\SetMathAlphabet{\mathsfit}{bold}{\encodingdefault}{\sfdefault}{bx}{n}

\usepackage{hyperref}
\hypersetup{hidelinks}
\usepackage{url}
\usepackage{booktabs}
\usepackage{amsmath}
\usepackage{amsthm}
\usepackage{graphicx}
\usepackage{xcolor}

\usepackage{algorithm}
\usepackage{algorithmic}

\newtheorem{proposition}{Proposition}

\title{Before Answering: Evidence Sufficiency under Size-Matched Memory Construction}

\author{\textbf{Joyanta Jyoti Mondal}\textsuperscript{1,*},
\textbf{Md. Shifatul Ahsan Apurba}\textsuperscript{2,*},
\textbf{Mridul Banik}\textsuperscript{3,*},
\\
\textbf{Md Masud Al Mahmud}\textsuperscript{4},
\textbf{Ibne Farabi Shihab}\textsuperscript{5}
\\
\textsuperscript{1}Department of Computer and Information Sciences, University of Delaware, USA\\
\textsuperscript{2}Department of Biomedical Informatics and Data Science,
University of Alabama at Birmingham, USA\\
\textsuperscript{3}Luddy School of
Informatics, Computing, and Engineering, Indiana University Indianapolis, USA\\
\textsuperscript{4}Department of Computer Science and Engineering, BRAC University, Bangladesh\\
\textsuperscript{5}Department of Computer Science, Iowa State University, USA\\
\small{
\textsuperscript{\textbf{*}}Equal Contribution.
\textbf{Correspondence:}
\href{mailto:joyanta@udel.edu}{joyanta@udel.edu}
}
}

\newcommand{\MemSafe}{\textsc{MemSafe}}
\newcommand{\LexSet}{\textsc{LexSet}}

\begin{document}
\maketitle

\begin{abstract}
Agents that answer questions from compressed or retrieved memory must recognize when the evidence a query needs is no longer in memory. Benchmarks for this task usually create insufficient-evidence examples by deleting supporting passages. We show that this construction leaks the label through memory size: on MuSiQue, a classifier that only counts paragraphs reaches an area under the ROC curve (AUROC) of $0.979$ for detecting unsafe memory, higher than the lexical estimator we initially evaluated. We propose a size-matched construction that provably removes this shortcut, and use it to study \MemSafe{}, an estimator that cross-encodes the query with each memory unit and aggregates the units with a set transformer. Across three multi-hop question answering datasets and five seeds, \MemSafe{} reaches $0.968$ and $0.983$ AUROC on MuSiQue and HotpotQA, $0.26$ to $0.39$ above a lexical baseline, while the third dataset, 2WikiMultiHopQA, is saturated. A frozen pretrained cross-encoder with a logistic head already closes $41\%$ of the MuSiQue gap between the lexical baseline and \MemSafe{}. At the same time, \MemSafe{} degrades more than a weak baseline on the unanswerable questions released with MuSiQue, reaches only $0.639$ AUROC on SQuAD~2.0, and needs several thousand clinical training examples before it outperforms a feature-based estimator. Used as a gate for a 7B reader, it reduces the error rate on answered questions from $0.850$ to $0.631$ at $5\%$ coverage, outperforming both reader confidence and, on average, the ground-truth integrity label, although a 7B LLM judge is the better gate at $10\%$ coverage. These results indicate that the way insufficient evidence is constructed matters as much as the estimator that detects it.
\end{abstract}

\section{Introduction}

\begin{figure}[t]
\centering
\includegraphics[width=0.99\linewidth]{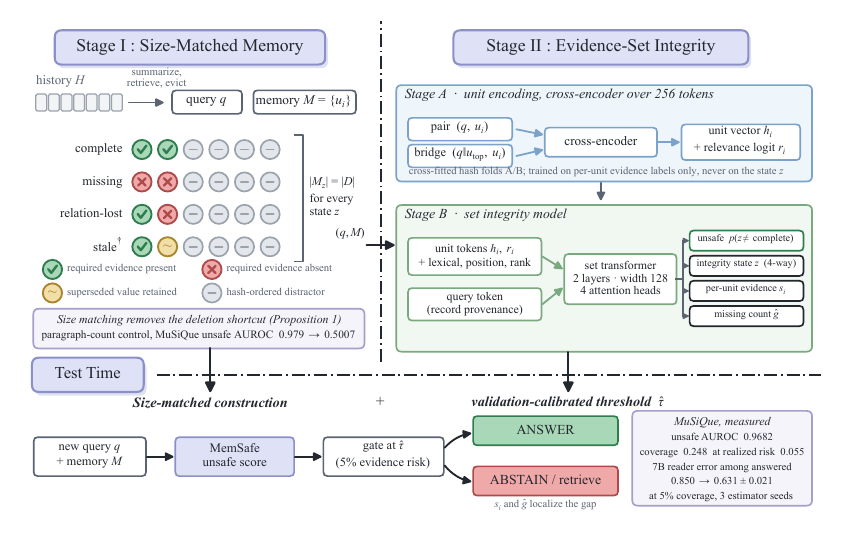}
\caption{Overview of \MemSafe{}. \textbf{Stage I} fixes the memory size at $|D|$ for every integrity state, so size carries no label information (Proposition~\ref{prop:size}). \textbf{Stage II} cross-encodes each $(q, u_i)$ pair and a bridge-conditioned pair $(q \Vert u_{\mathrm{top}}, u_i)$ (Stage A), and a two-layer set transformer (Stage B) predicts the unsafe probability, the integrity state, per-unit evidence presence $s_i$, and the number of missing evidence items $\widehat{g}$. At \textbf{test time}, a threshold $\widehat{\tau}$ chosen on validation data at a $5\%$ risk budget decides whether to answer. Values are MuSiQue test-split means.}
\label{fig:overview}
\end{figure}

Language-model agents that operate over long horizons cannot keep their full interaction history in context. They summarize, retrieve, and evict, and the remaining memory may no longer contain the evidence a query depends on, although an answer produced from it looks no different. Memory benchmarks show that such agents struggle to retain and use information over long interactions \citep{tan-etal-2025-membench,jia-etal-2025-evaluating}, that memory addition and deletion choices shape their later behavior \citep{xiong-etal-2026-memory}, and that longer inputs alone can degrade performance even when the relevant evidence is retrieved \citep{du-etal-2025-context}. Compressing context to cope with this length can itself discard what a question needs \citep{yoon-etal-2024-compact}. The problem is especially costly in clinical settings, where agents increasingly reason over electronic health records \citep{shi-etal-2024-ehragent} and a record can retain an outdated measurement after its update has been lost. Before answering, an agent therefore needs to estimate whether its memory is sufficient and accurate for the query.

Recent work treats sufficiency as a learnable signal. \citet{joren2025sufficient} introduce sufficient context as a criterion for selective generation. \citet{li2026s2grag} predict sufficiency and structured evidence gaps to guide retrieval, and \citet{qiu2026surerag} verify sufficiency over sets of passages. Evaluating such estimators requires examples with missing evidence, usually created by deleting supporting passages from a sufficient context, and \citet{qiu2026evidence} show that labels built this way contain exploitable artifacts.

We find that the resulting shortcut can be severe. In our initial construction, supporting paragraphs are removed without replacement, so complete memories always contain more paragraphs than corrupted ones. As a result, a classifier that only counts paragraphs reaches $0.979$ unsafe AUROC on MuSiQue, which is higher than the $0.968$ of the lexical estimator we initially set out to evaluate. Once memory size is held fixed across integrity states, the same classifier performs at chance.

In this work, we ask what an estimator can still detect once this shortcut is removed, and which components are responsible. We study \MemSafe{} (Figure~\ref{fig:overview}), which scores each (query, memory unit) pair with a cross-encoder and aggregates the units with a small set transformer. We compare it with the lexical estimator it replaces (\LexSet{}), matched comparators, long-context joint encoders, a large language model (LLM) judge, and surface-level controls. Our contributions are as follows:
\begin{itemize}
  \item \textbf{Size-matched construction.} We identify a memory-size leak in deletion-based benchmarks and propose a construction under which every size-only estimator has chance-level AUROC when each base query contributes the same integrity states, and a bounded deviation otherwise. 
  \item \textbf{Attribution.} We separate the contributions of the pretrained cross-encoder, fine-tuning, set-level interaction, and gold versus silver supervision. 
  \item \textbf{Stress tests.} We evaluate the estimator on a second construction, on SQuAD~2.0, and on clinical records at several scales, and report where it underperforms. 
  \item \textbf{Downstream gating.} We show, empirically and formally, that a continuous sufficiency score can be a better answer gate than the ground-truth integrity label. 
\end{itemize}

\section{Related Work}

\paragraph{Context sufficiency and evidence verification.} \citet{joren2025sufficient} distinguish whether a context is sufficient to answer a query from whether a model answers it correctly, and combine a sufficiency signal with model confidence for selective generation. \citet{li2026s2grag} use a judge to predict sufficiency and structured gaps that drive follow-up retrieval. \citet{qiu2026surerag} instead label a set of passages as supporting, refuting, or not establishing a candidate answer. Our setting follows the answer-free formulation of \citet{joren2025sufficient}, which judges sufficiency from the query and context alone. We differ in replacing an LLM autorater with a compact supervised estimator that also identifies which memory units carry the required evidence, and in controlling how insufficient examples are constructed.

\paragraph{Construction artifacts.} \citet{qiu2026evidence} study not-enough-information (NEI) labels created by deleting evidence and show that models can exploit artifacts of the deletion instead of assessing sufficiency. We remove one such artifact, memory size, by construction, and test whether estimators trained on one construction transfer to another (Section~\ref{sec:transfer}).

\paragraph{Multi-hop evidence and retrieval control.} Multi-hop question answering (QA) datasets annotate the evidence each question requires \citep{yang2018hotpotqa,ho2020constructing,trivedi2022musique}. \citet{zhang2026failure} find that multi-hop accuracy is limited by the least visible piece of evidence and also depends on its absolute position. Adaptive retrieval decides when and how much to retrieve: methods route queries on predicted complexity \citep{jeong2024adaptiverag}, interleave retrieval with chain-of-thought reasoning \citep{trivedi2023ircot}, or escalate retrieval by recoverability and cost \citep{li2026raser}; Appendix~\ref{app:related} reviews further retrieval gates, conflict handling, and unanswerability detection. These gates ask whether the model needs more information, whereas our estimator asks whether the memory it already holds contains the evidence. It is complementary: it predicts from the memory alone whether it is sufficient and which evidence is missing, information that such controllers can use.

\paragraph{Selective prediction.} Selective classification formalizes when a model should abstain \citep{chow1970recognition,geifman2017selective,geifman2019selectivenet}. This is typically done by thresholding a confidence score \citep{hendrycks2017baseline}, whose calibration can be measured directly \citep{guo2017calibration}. These methods, and the extensions and LLM abstention work reviewed in Appendix~\ref{app:related}, threshold scores derived from the model's own predictions; distribution-free risk control calibrates such thresholds with finite-sample guarantees \citep{angelopoulos2022risk,angelopoulos2024learn}. We instead study a score derived from the evidence, choose its threshold on validation data without such a guarantee, and evaluate it as an answer gate at fixed coverage (Section~\ref{sec:reader}).

\section{Method}
\label{sec:method}

\subsection{Problem Formulation}

Let $H$ denote an interaction history, $M$ the memory an agent retains from it, $q$ a query, and $E(q)$ the set of evidence items required to answer $q$. The memory consists of units $u_1,\dots,u_n$, each with observable provenance (position, rank, and record size). We describe the relation between $M$ and $E(q)$ by an integrity state $z$ with four values: \textsc{complete} if every item of $E(q)$ is present and current, \textsc{missing} if none is present, \textsc{relation-lost} if some but not all are present, and \textsc{stale} if a superseded value is present while its update is absent. \textsc{stale} requires temporal supersession and occurs only in the clinical data. We call a memory \emph{unsafe} if $z \neq \textsc{complete}$; Appendix~\ref{app:notation} collects the notation. An estimator predicts $p(z \mid q, M)$ and the unsafe probability $p(z \neq \textsc{complete} \mid q, M)$, which a downstream policy can threshold. To support targeted retrieval, it also predicts the probability $s_i = p(u_i \in E(q) \mid q, M)$ that each unit carries required evidence and an estimate $\widehat{g}$ of the number of missing items $g = |E(q) \setminus M|$. In the QA data, $E(q)$ is the annotated set of supporting paragraphs. Such annotations need not enumerate every sufficient evidence set: removing an annotated unit need not remove every route to the answer, and retaining all of them need not make a reader correct.

A gate answers when the unsafe score $r(q,M)$ is at most a threshold $t$, which defines its coverage $C(t)=\Pr(r \le t)$. We distinguish two risks among accepted examples: the \emph{evidence risk} $R_E(t)=\Pr(z \neq \textsc{complete} \mid r \le t)$ and the \emph{reader risk} $R_Y(t)=\Pr(Y=1 \mid r \le t)$, where $Y$ indicates that a reader answers incorrectly. The integrity label defines $R_E$, answer correctness defines $R_Y$, and controlling one does not bound the other.

\subsection{Size-Matched Memory Construction}
\label{sec:construction}

Let $S \subseteq M$ be the units that carry required evidence and $D = M \setminus S$ the remaining units. A deletion-based construction forms a \textsc{missing} memory as $D$ and a \textsc{complete} memory as $S \cup D$. Then $|M_{\textsc{missing}}| < |M_{\textsc{complete}}|$, and the memory size alone identifies $z$. We instead give every state the same size, $|M_z| = |D|$: a state that retains $k$ evidence units contains those units and $|D| - k$ distractors, selected in a fixed hash order. Base queries with $|D| < |S|$ cannot be matched and are discarded. Memory size and compression ratio are then constant within each base query.

\begin{proposition}[Size invariance]\label{prop:size}
Let $f$ be any estimator whose score depends on the memory only through its size, including the compression ratio $|M|/|H|$, so that under the size-matched construction it assigns one score to all variants of a base query. Let $P_u$ and $P_c$ be the distributions of the base query of a uniformly drawn unsafe and complete example. Then $|\mathrm{AUROC}(f) - 1/2| \le \mathrm{TV}(P_u, P_c)$. In particular, if every base query contributes the same numbers of unsafe and complete examples, $f$ has unsafe AUROC exactly $1/2$.
\end{proposition}

Examples from the same base query receive tied scores, and cross-query pairs cancel when $P_u = P_c$. The bound depends on how many examples the exceptional queries contribute, not only on how many such queries there are: a single query with many unsafe variants can dominate $P_u$. We give the proof, a counterexample to a bound in terms of the fraction of exceptional queries, and the exact MuSiQue calculation in Appendix~\ref{app:proofs}, and test the prediction in Section~\ref{sec:validation}.

\subsection{Two-Stage Estimator}

\paragraph{Stage A: unit encoding.} A pretrained MiniLM cross-encoder \citep{wang2020minilm} encodes each pair $(q, u_i)$, truncated to 256 tokens, into a unit representation and a relevance logit. In contrast to a bi-encoder \citep{reimers2019sentencebert}, which embeds the query and the unit separately, a cross-encoder lets each query token attend to each unit token. Second-hop evidence is often lexically unrelated to the query, so we add a bridge-conditioned encoding of $(q \,\|\, u_{\text{top}},\, u_i)$, where $u_{\text{top}}$ is the highest-scoring unit under the pretrained encoder.

\paragraph{Encoder fine-tuning.} We fine-tune the encoder for one epoch to predict whether each unit carries required evidence, an objective that never uses the integrity label. To keep Stage B from training on embeddings of units the encoder was fine-tuned on, each of two hash-defined training folds is embedded by the encoder fine-tuned on the other fold, while validation and test units are embedded by the average of both encoders. Training and evaluation embeddings therefore come from slightly different encoders (one versus the average of two).

\paragraph{Stage B: set integrity model.} Each unit token concatenates the two encodings, the two relevance logits, and the lexical similarity, position, and rank features of \LexSet{}; a learned query token carries record-level provenance. Two transformer encoder layers let the units attend to one another, as in set transformers \citep{lee2019set}. The query token feeds the unsafe, integrity-state, and missing-count heads, and each unit token feeds the per-unit evidence head. The loss sums binary cross-entropy for the unsafe label, cross-entropy over states, masked per-unit binary cross-entropy, and a smooth $L_1$ loss for the missing count. Because the unsafe and integrity-state heads are separate, the unsafe probability need not equal one minus the predicted probability of \textsc{complete}.

\paragraph{Matched comparators.} To isolate the set model, two comparators use identical embeddings: mean pooling with the same heads but no attention between units, and a logistic model over aggregated relevance scores (maximum, top-$k$ means, and counts above a threshold). We also evaluate a cross-encoder over the concatenated memory, the strongest baseline of \citet{qiu2026surerag}.

\section{Experimental Setup}
\label{sec:setup}

\paragraph{Data.} We use the full processed MuSiQue \citep{trivedi2022musique}, HotpotQA \citep{yang2018hotpotqa} (distractor setting), and 2WikiMultiHopQA \citep{ho2020constructing} datasets, with $21{,}902$, $97{,}406$, and $180{,}030$ base questions that yield $65{,}297$, $292{,}218$, and $540{,}090$ memory variants. Every excluded record has a recorded reason, such as too few distractors or missing supporting facts. Family A is the size-matched construction of Section~\ref{sec:construction}. Family B consists of the unanswerable questions released with MuSiQue, which remove the answer paragraph of one sub-question at equal context length. Both families share one split function, so no test question in one family appears in training in the other.

\paragraph{Training and evaluation.} Splits are assigned by a salted hash of the base question before corruption. We use five pre-specified seeds, $\{17,29,43,71,101\}$. Encoder fine-tuning and the concatenated cross-encoder use the first three; Stage B seeds $71$ and $101$ reuse the encoders of seeds $17$ and $29$, so the five seeds are not five independent repetitions of the full pipeline. The Stage B configuration (width 128, dropout 0.1, learning rate $3\times10^{-4}$) is selected by MuSiQue validation AUROC from a pre-specified eight-point grid, whose scores all fall between $0.963$ and $0.968$. Our primary metric is unsafe AUROC, the AUROC for separating unsafe from complete memories. We also report integrity macro F1, expected calibration error (ECE), the area under the precision-recall curve (AUPRC) for per-unit evidence, and the mean absolute error (MAE) of the missing count. Confidence intervals come from a paired hierarchical bootstrap with $1{,}000$ resamples over base questions and seeds; they do not include a separate level for encoder training, so they understate full-pipeline variation. With five seeds, a Wilcoxon signed-rank test cannot reach $p<0.05$ (the smallest attainable two-sided $p$ is $0.0625$), so we report it only descriptively.

\paragraph{Pre-registered hypotheses.} Before the final runs, we register that \MemSafe{} beats mean pooling (H1), beats \LexSet{} and a flat term frequency-inverse document frequency (TF-IDF) classifier (H2), degrades less than TF-IDF and text length from family A to B (H3), gains from bridge conditioning on MuSiQue and 2WikiMultiHopQA (H4), and beats \LexSet{} on MIMIC-IV~3.1 (H5); Appendix~\ref{app:hypotheses} gives the criteria.

\section{Results}

\subsection{Size Matching Removes the Shortcut}
\label{sec:validation}

\begin{figure}[t]
\centering
\includegraphics[width=0.99\linewidth]{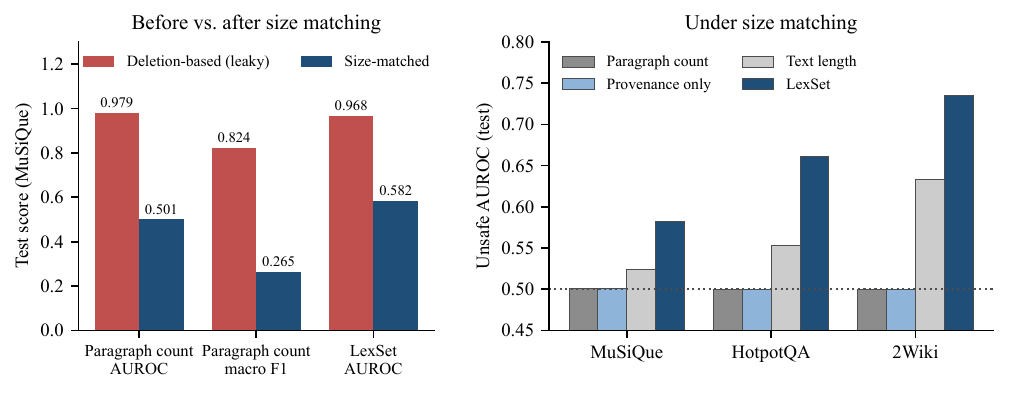}
\caption{Size matching removes the memory-size shortcut. Left: MuSiQue test scores under the deletion construction (taken from the pre-fix run ledger) and under size matching. Right: unsafe AUROC of the surface controls and \LexSet{} under size matching. The dotted line marks chance.}
\label{fig:shortcut}
\end{figure}

Figure~\ref{fig:shortcut} compares the surface controls before and after size matching. Under the deletion construction, the paragraph-count control reaches $0.979$ unsafe AUROC on MuSiQue, above the $0.968$ that \LexSet{}, the lexical estimator we initially evaluated, reaches on the same construction. After size matching, the control drops to $0.5007$ on MuSiQue and to exactly $0.5000$ on HotpotQA and 2WikiMultiHopQA, as Proposition~\ref{prop:size} predicts: every HotpotQA and 2WikiMultiHopQA base question contributes all three states. Every MuSiQue base question keeps its complete and missing variants, and $1.87\%$ lack the relation-lost variant, which bounds the deviation by $0.0046$ over the processed dataset (Appendix~\ref{app:proofs}), against an observed $0.0007$. Because distractors are added in a fixed hash order, unit positions could still depend on the state, but a control that sees only positional statistics also scores $0.5007$, $0.5000$, and $0.5000$ (Table~\ref{tab:main}). Size matching does not remove every surface cue: text length still reaches $0.525$ to $0.633$ AUROC, highest on 2WikiMultiHopQA, against $0.515$ on the SQuAD~2.0 development set, which we did not construct. The construction is therefore count-matched rather than free of artifacts. All remaining results use the size-matched construction.

\subsection{Main Results}
\label{sec:main}

Table~\ref{tab:main} reports unsafe AUROC and integrity macro F1 on the test split. \MemSafe{} achieves the best score on both metrics on all three datasets and outperforms every baseline on each of the five seeds, which supports H2. Its paired bootstrap advantage over \LexSet{} is $+0.3862$ $[+0.3790,+0.3933]$ on MuSiQue, $+0.3222$ $[+0.3191,+0.3257]$ on HotpotQA, and $+0.2648$ $[+0.2579,+0.2691]$ on 2WikiMultiHopQA, and its advantage over flat TF-IDF ($+0.2636$, $+0.3614$, and $+0.1298$) also has intervals that exclude zero. \MemSafe{} localizes evidence more accurately than the mean-pooling comparator (per-unit AUPRC of $0.9123$ versus $0.8850$ on MuSiQue) and is well calibrated, with ECE of at most $0.0128$. On 2WikiMultiHopQA, all encoder-based models exceed $0.99$ AUROC; we return to this saturation in Section~\ref{sec:transfer}.

\begin{table}[t]
\caption{Test-split unsafe AUROC and integrity macro F1, averaged over five seeds (three for the concatenated and 8k joint encoders). Seed-level standard deviations of the learned models are at most $0.0031$; deterministic models have none. n/a: not run. Bold marks the best value in each column; $\uparrow$ means higher is better.}
\label{tab:main}
\small
\begin{center}
\resizebox{\linewidth}{!}{%
\begin{tabular}{lcccccc}
\toprule
& \multicolumn{2}{c}{MuSiQue} & \multicolumn{2}{c}{HotpotQA} & \multicolumn{2}{c}{2WikiMultiHopQA} \\
\cmidrule(lr){2-3}\cmidrule(lr){4-5}\cmidrule(lr){6-7}
Model & AUROC $\uparrow$ & Macro F1 $\uparrow$ & AUROC $\uparrow$ & Macro F1 $\uparrow$ & AUROC $\uparrow$ & Macro F1 $\uparrow$ \\
\midrule
\MemSafe{} (ours) & $\mathbf{0.9682}$ & $\mathbf{0.7960}$ & $\mathbf{0.9831}$ & $\mathbf{0.8600}$ & $\mathbf{0.9999}$ & $\mathbf{0.9894}$ \\
Encoder + mean pool & $0.9484$ & $0.7665$ & $0.9765$ & $0.8423$ & $0.9997$ & $0.9880$ \\
Relevance aggregation & $0.9215$ & $0.7136$ & $0.9612$ & $0.8146$ & $0.9947$ & $0.9758$ \\
Concatenated cross-encoder & $0.8912$ & $0.5811$ & $0.9590$ & $0.7575$ & $0.9983$ & $0.9579$ \\
8k joint encoder (three seeds; exploratory) & $0.9019$ & $0.5974$ & $0.9559$ & $0.7534$ & n/a & n/a \\
Flat TF-IDF logistic & $0.7046$ & $0.4408$ & $0.6217$ & $0.3902$ & $0.8702$ & $0.6558$ \\
Sufficient-context-style (TF-IDF) & $0.7022$ & $0.3700$ & $0.6228$ & $0.3287$ & $0.8730$ & $0.4531$ \\
\LexSet{} lexical event-set & $0.5820$ & $0.3111$ & $0.6609$ & $0.3916$ & $0.7351$ & $0.5029$ \\
Retrieval-similarity proxy & $0.5354$ & $0.2986$ & $0.5903$ & $0.3544$ & $0.5446$ & $0.3065$ \\
Text-length control & $0.5246$ & $0.3045$ & $0.5532$ & $0.3159$ & $0.6328$ & $0.3772$ \\
Paragraph-count control & $0.5007$ & $0.2645$ & $0.5000$ & $0.1667$ & $0.5000$ & $0.1667$ \\
Provenance-only control & $0.5007$ & $0.2366$ & $0.5000$ & $0.1667$ & $0.5000$ & $0.1667$ \\
Majority & $0.5000$ & $0.1675$ & $0.5000$ & $0.1667$ & $0.5000$ & $0.1667$ \\
\bottomrule
\end{tabular}
}
\end{center}
\end{table}

\paragraph{Long-context and LLM baselines.} A ModernBERT joint encoder \citep{warner2025smarter} with an $8{,}192$-token window, which truncates no MuSiQue or HotpotQA memory, reaches only $0.9019$ and $0.9559$ AUROC, and a Qwen2.5-7B-Instruct \citep{qwen2025qwen25technicalreport} sufficiency judge in the style of \citet{joren2025sufficient} reaches $0.7832$ zero-shot and $0.7825$ four-shot on a $1{,}000$-question MuSiQue subset, against $0.9719$ for \MemSafe{} (Appendix~\ref{app:hardening}).

\paragraph{Transfer to an independent benchmark.} Applied without adaptation to the SQuAD~2.0 development set \citep{rajpurkar2018squad2}, whose unanswerable questions are written by crowdworkers rather than produced by deletion, the MuSiQue-trained model reaches $0.6393$ AUROC. This is above flat TF-IDF ($0.5099$) but far below its in-distribution performance, and its MuSiQue threshold for a $5\%$ risk budget accepts only $0.0049$ of questions yet realizes a risk of $0.2759$, so the transferred threshold does not control risk.

\subsection{Sources of the Improvement}
\label{sec:component-ablations}

When \MemSafe{} is built up from \LexSet{} in stages, most of the improvement comes from replacing lexical similarity with a fine-tuned cross-encoder, before any set model is added: AUROC rises from $0.582$, $0.661$, and $0.735$ to $0.922$, $0.961$, and $0.995$. Mean pooling and inter-unit attention add a further $0.046$, $0.022$, and $0.005$, and attention yields a small but consistent gain over mean pooling (H1): $+0.0199$ on MuSiQue (interval $[+0.0165,+0.0242]$), $+0.0067$ on HotpotQA ($[+0.0058,+0.0076]$), and $+0.0002$ on 2WikiMultiHopQA ($[+0.0001,+0.0004]$). H1 therefore holds, but the gain shrinks as the task becomes easier, and under family-B transfer the macro-F1 difference is not significant ($-0.0010$, interval $[-0.0151,+0.0167]$).

\begin{table}[t]
\caption{Single-component ablations on MuSiQue (test split, three Stage B seeds sharing one encoder). Each row removes one component from the full model, except the last, which removes both fine-tuning and the set model. Bold marks the best value in each column; $\uparrow$ means higher is better.}
\label{tab:ablation}
\small
\begin{center}
\begin{tabular}{lcc}
\toprule
Variant & Unsafe AUROC $\uparrow$ ($\Delta$) & Macro F1 $\uparrow$ ($\Delta$) \\
\midrule
Full \MemSafe{} & $0.9674 \pm 0.0008$ & $\mathbf{0.7929} \pm 0.0022$ \\
\quad w/o encoder fine-tuning (frozen encoder) & $0.9121$ ($-0.0553$) & $0.6333$ ($-0.1596$) \\
\quad w/o bridge-conditioned encoding & $0.9423$ ($-0.0251$) & $0.7551$ ($-0.0377$) \\
\quad w/o set model (mean pooling) & $0.9502$ ($-0.0173$) & $0.7664$ ($-0.0265$) \\
\quad w/o lexical and position features & $0.9637$ ($-0.0038$) & $0.7839$ ($-0.0090$) \\
\quad w/o localization heads & $\mathbf{0.9677}$ ($+0.0002$) & $0.7915$ ($-0.0014$) \\
\quad w/o encoder fine-tuning \emph{and} set model & $0.7407$ ($-0.2268$) & $0.4602$ ($-0.3327$) \\
\bottomrule
\end{tabular}
\end{center}
\end{table}

\paragraph{Ablations.} In Table~\ref{tab:ablation}, freezing the encoder causes the largest drop, followed by removing the bridge-conditioned encoding, which supports H4 on MuSiQue (the ablation is not run on 2WikiMultiHopQA), and by replacing the set model with mean pooling. The lexical and positional features and the localization heads, which exist for per-unit outputs, barely affect detection. Without both fine-tuning and the set model, the model still reaches $0.7407$, so a frozen pretrained cross-encoder already closes $0.1587$ of the $0.3854$ gap between \LexSet{} and the full model ($41\%$). This is a descriptive position on the ladder, not a causal allocation of credit among components. Fine-tuning and the set model largely overlap (Figure~\ref{fig:interaction}, Appendix~\ref{app:ablation}): each adds $0.17$ to $0.18$ AUROC when applied alone, but only $0.04$ to $0.06$ when added on top of the other. Because this comparison mixes two replication scopes, it shows diminishing returns rather than a matched interaction estimate.

\paragraph{Supervision without gold unit targets.} Replacing the per-unit training targets with silver labels from answer-string matching lowers AUROC only from $0.9674$ to $0.9444$, but per-unit AUPRC from $0.9123$ to $0.4252$; gold supporting facts still define the variants and integrity labels (Appendix~\ref{app:silver}).

\subsection{Transfer Across Constructions and Saturation}
\label{sec:transfer}

Trained on family A and evaluated on family B, \MemSafe{} remains far ahead in absolute terms ($0.8948$ AUROC against $0.6496$ for flat TF-IDF) but loses more under the shift: $0.0734$ AUROC, against $0.0550$ for flat TF-IDF. H3 is therefore refuted; the text-length control does not degrade, but it is near chance on both families. The ECE of \MemSafe{} also rises from $0.0128$ to $0.071$, above the $0.038$ of mean pooling. As \citet{qiu2026evidence} observe, robustness to one construction of insufficiency does not carry over to another.

\paragraph{Memorization and saturation.} Restricting evaluation to questions whose evidence never appears in training changes \MemSafe{} by at most $0.019$ AUROC, while flat TF-IDF loses up to $0.153$ (Appendix~\ref{app:memorization}). The near-perfect 2WikiMultiHopQA scores ($0.9999$ AUROC) more likely reflect templated questions that name the entities whose pages contain the evidence, so we treat this dataset as saturated and rely on MuSiQue to compare strong models.

\subsection{Selective Prediction at a Fixed Risk Budget}
\label{sec:selective}

A deployed gate operates at a single threshold chosen in advance. We choose, on validation data, the threshold at which the evidence risk $R_E$ (the fraction of accepted memories that are not complete) is $5\%$, apply it to the test set, and report the resulting coverage. Prevalence caps this coverage: only one third of QA variants are complete, so even a perfect ranker accepts at most $(1/3)/(1-0.05) \approx 0.3509$ at a $5\%$ evidence risk (Appendix~\ref{app:selective}). At this operating point, models differ far more than their AUROC suggests. On MuSiQue, \MemSafe{} and mean pooling differ by only $0.0199$ AUROC, yet \MemSafe{} accepts $1.86$ times as many memories ($0.2479$ versus $0.1334$ coverage). Their realized test risks differ ($0.0553$ versus $0.0463$), so this compares the two at a shared validation target rather than at equal test risk. On 2WikiMultiHopQA, \MemSafe{} ($0.3509$) and mean pooling ($0.3502$) both sit at this ceiling, so the saturated dataset cannot separate them at this operating point. For \MemSafe{}, the realized test risk stays between $0.0502$ and $0.0553$ on the three datasets, whereas the validation threshold of \LexSet{} yields a test risk of $0.7333$ at almost zero coverage. Appendix~\ref{app:selective} reports the clinical operating points and the error taxonomy.

\subsection{Gating a Reader}
\label{sec:reader}

We next ask whether the sufficiency score reduces wrong answers when it decides which questions a reader answers. Qwen2.5-1.5B-Instruct and Qwen2.5-7B-Instruct answer from each MuSiQue memory variant with greedy decoding, on fixed samples of $1{,}500$ and $1{,}000$ base questions, scored by normalized exact match; a gate answers only the memories with the lowest unsafe scores. Even with complete memory, exact match is only $0.156$ and $0.291$, so answering everything is wrong $91.6\%$ and $85.0\%$ of the time and none of the gates we evaluate meets our pre-registered reader-risk targets of $5\%$ to $20\%$. We therefore report the reader risk $R_Y$ at fixed coverage, a protocol we adopt after observing the base error rate, and treat this analysis as exploratory.

For the 7B reader, averaged over three estimator seeds (Figure~\ref{fig:clinicalreader}, right), gating on \MemSafe{} lowers the error on answered questions from $0.8503$ to $0.6306 \pm 0.0206$ at $5\%$ coverage; the 1.5B reader shows the same trend ($0.8296$ against $0.9162$ for answering everything, at $10\%$ coverage). The gate outperforms the reader's sequence confidence at every coverage level and seed, and for the 1.5B reader, confidence gating is even worse than answering everything ($0.9731$), so answer confidence does not capture evidence sufficiency. On average, gating on \MemSafe{} also outperforms gating on the ground-truth integrity label at all four coverage levels, with a single near-tie for one seed at $25\%$ coverage ($0.7160$ versus $0.7133$).

Proposition~\ref{prop:coarse} (Appendix~\ref{app:proofs}) formalizes why: at fixed coverage, reader risk is minimized by ranking on the reader's error probability $\eta(q,M)=\Pr(Y=1\mid q,M)$, and a gate that ranks by a binary variable such as the integrity label attains this minimum at every coverage only if $Y$ is conditionally independent of $(q,M)$ given that variable. Complete memories differ in how difficult their evidence is to use, so a score that tracks the reader's error probability can beat a binary label that ignores this variation. The LLM judge, from the reader's own model family, is the better gate at $10\%$ coverage, while \MemSafe{} is better at $25\%$ and $50\%$ (Table~\ref{tab:reader}). This is consistent with the same reasoning, although our experiments do not isolate a model-family effect. The ordering also holds at lower unsafe prevalence in a single-seed analysis (Appendix~\ref{app:reader}). Still, about two thirds of answered questions remain wrong at $5\%$ coverage, so the gate improves which questions are answered without yielding a deployable operating point on MuSiQue.

\subsection{Clinical Evaluation}
\label{sec:clinical}

\begin{figure}[t]
\centering
\includegraphics[width=0.99\linewidth]{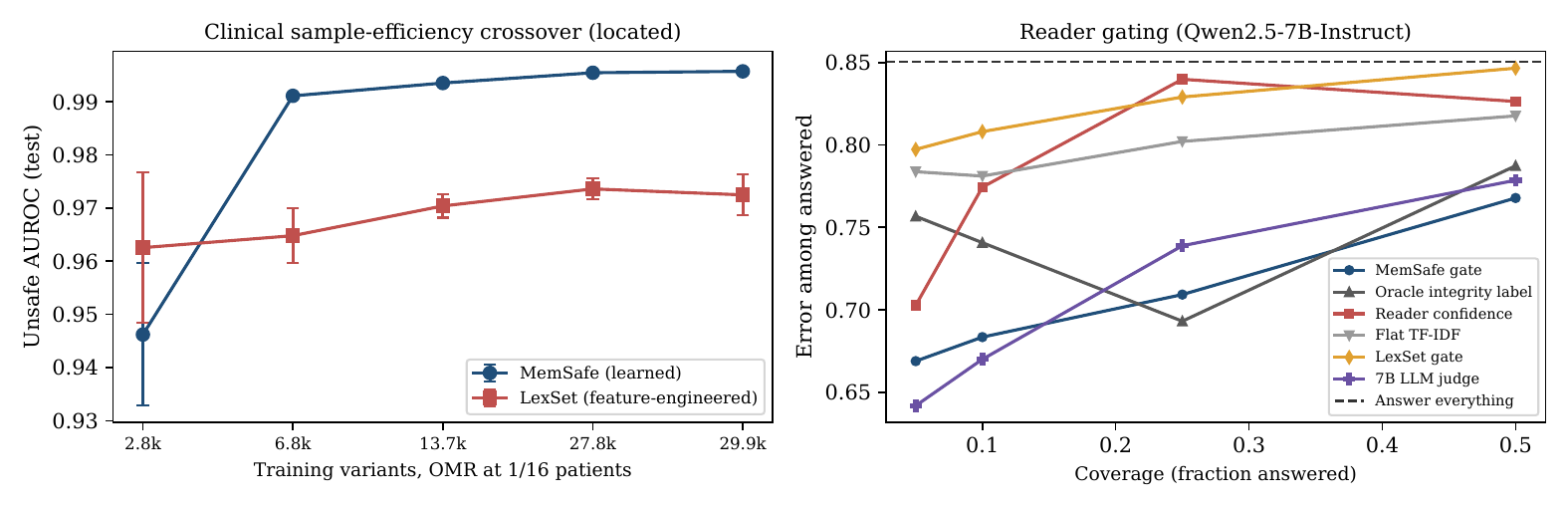}
\caption{Left: clinical learning curve on MIMIC-IV OMR at a fixed event source and patient density ($1/16$ of patients, three seeds per point). \LexSet{} leads at the smallest scale, and \MemSafe{} overtakes it with about $2.5$ times as many training variants; realized patient counts also grow with the cap. Right: selective answering with the 7B reader. Gating on the predicted unsafe score yields lower error than answering everything, than reader confidence, and, on average over three estimator seeds, than the ground-truth integrity label at every coverage level shown; the LLM judge is the better gate at $10\%$ coverage.}
\label{fig:clinicalreader}
\end{figure}

Clinical memories hold numeric measurements whose integrity depends on temporal order, and they are the only setting with \textsc{stale}. We use two MIMIC-IV~3.1 sources \citep{johnson2023mimiciv}, OMR and laboratory events, with five seeds and deterministic patient sampling. In a small configuration that reproduces the prior lexical system ($1{,}000$ base questions, $683$ patients), \LexSet{} ranks memories better than \MemSafe{} ($0.9649$ versus $0.9220$ AUROC), although \MemSafe{} predicts the state more accurately. A learning curve on OMR that varies only the question cap argues against a modality explanation (Figure~\ref{fig:clinicalreader}, left), although patient counts also grow along it ($929$ to $3{,}934$). \LexSet{} still leads at $2{,}788$ training variants ($0.9626$ versus $0.9462$), but \MemSafe{} leads on all three seeds from $6{,}844$ variants onward ($0.9911$ versus $0.9648$). Laboratory events show the same ordering at the largest scale (Appendix~\ref{app:clinical}), so H5 is only partially supported. Because \LexSet{} encodes numeric trends as explicit features, it remains the better ranker for small clinical cohorts.

\section{Limitations}
\label{sec:limitations}

The memories we evaluate are built from QA benchmarks and clinical records, not produced by deployed agents. The main models use gold supporting-fact labels; silver unit targets preserve detection but not localization, and gold annotations still define the memory variants and integrity labels, so learning from memories without annotated evidence remains open. The sufficient-context-style baseline approximates the autorater of \citet{joren2025sufficient} with TF-IDF, and the LLM judge uses one pinned model with fixed prompts. Inference rests on paired bootstrap intervals, and because we re-derive the splits, our numbers are not directly comparable with published development-set results. The clinical learning curve covers one event source with three seeds per point, and the reader study covers one dataset with greedy decoding. Finally, we evaluate the score as an answer gate; we leave retrieval and abstention policies that act on it to future work.

\section{Discussion and Conclusion}

We study how an agent can estimate, before answering, whether its memory still contains the evidence a query requires. Deletion-based construction lets a trivial size control outperform the estimator we initially evaluated. Once size is matched, most of the signal comes from a pretrained cross-encoder, and fine-tuning and set-level interaction add overlapping gains, which qualifies the emphasis on set-level verification in \citet{qiu2026surerag} for the answer-free setting. The resulting estimator is a better answer gate than reader confidence and, on average, the ground-truth integrity label, but the strongest in-construction model is not the most robust, and a small clinical study would favor the lexical estimator. We therefore recommend evaluating sufficiency estimators against size-matched controls, on a held-out construction family, and at the data scale of the intended deployment. Future work should localize missing evidence without gold annotations and combine answer-free sufficiency estimates with reader-aware judges.

\section*{AI Use Statement}
Generative AI tools assist with language editing, literature discovery, mathematical exposition, and source-code preparation. The authors verify the proofs, citations, experimental records, and final claims and remain responsible for the manuscript.

\subsection*{Ethics Statement}
The clinical component of this project uses de-identified retrospective data under a credentialed data use agreement \citep{johnson2023mimiciv}; no result here supports clinical deployment, improved patient outcomes, or clinician-level performance. The evidence-QA experiments use public research benchmarks. Corrupted-memory states are constructed artifacts, not observed clinical failures.

\subsection*{Reproducibility Statement}
The appendix contains complete proofs of both propositions, the dataset construction and exclusion counts, the training configuration and selection grid, and the pre-registered criteria with their outcomes. The accompanying code records raw and processed-data checksums, per-seed metrics, and a fingerprint of the code, data, and settings of every run separately from the aggregated tables reported here, and a verification script re-checks the chain from raw files to run coverage. MIMIC-IV records are not redistributed and require credentialed access.

\bibliography{iclr2027_conference}
\bibliographystyle{iclr2027_conference}

\appendix

\section{Overview and Notation}
\label{app:notation}

The appendix is organized as follows. Appendix~\ref{app:related} extends the related work. Appendix~\ref{app:proofs} proves Propositions~\ref{prop:size} and~\ref{prop:coarse}. Appendix~\ref{app:experimental} describes the datasets, training, and pre-registered criteria. Appendix~\ref{app:results} reports additional results. Table~\ref{tab:notation} summarizes our notation.

\begin{table}[h]
\caption{Notation.}
\label{tab:notation}
\small
\begin{center}
\begin{tabular}{ll}
\toprule
Symbol & Meaning \\
\midrule
$H$ & Interaction history \\
$M$, $M_z$ & Memory retained from $H$; memory constructed for integrity state $z$ \\
$q$ & Query \\
$E(q)$ & Evidence items required to answer $q$ \\
$u_i$, $n$ & The $i$-th memory unit; number of units in $M$ \\
$z$ & Integrity state: \textsc{complete}, \textsc{missing}, \textsc{relation-lost}, or \textsc{stale} \\
$S$, $D$ & Units that carry required evidence; remaining (distractor) units \\
$k$ & Number of evidence units retained in a constructed state \\
$s_i$ & Predicted probability that $u_i$ carries required evidence \\
$\widehat{g}$ & Predicted number of missing evidence items \\
$u_{\text{top}}$ & Highest-scoring unit under the pretrained encoder \\
$f$ & Estimator that depends on the memory only through its size (Proposition~\ref{prop:size}) \\
$P_u$, $P_c$ & Distributions of the base query of a uniformly drawn unsafe and complete example \\
$\widehat{\tau}$ & Answer threshold selected on validation data \\
$r(q,M)$, $t$ & Predicted unsafe score; acceptance threshold \\
$c$, $C(t)$ & Coverage, the fraction of examples a gate accepts \\
$R_E(t)$ & Evidence risk: fraction of accepted memories that are not complete \\
$R_Y(t)$ & Reader risk: fraction of accepted examples the reader answers incorrectly \\
$\pi$ & Unsafe prevalence, $\Pr(z \neq \textsc{complete})$ \\
$Y$ & Indicator that the reader answers incorrectly \\
$\eta(q,M)$ & Probability that the reader answers incorrectly given $(q, M)$ \\
$Z$ & Binary variable used by a coarse gate, such as the ground-truth integrity state \\
$b$, $D_b$ & Base query and its distractor set (proofs) \\
$n_u$, $n_c$ & Numbers of unsafe and complete examples contributed by a base query (proofs) \\
\bottomrule
\end{tabular}
\end{center}
\end{table}

\section{Extended Related Work}
\label{app:related}

\paragraph{Retrieval gates and related controllers.} Beyond routing by complexity, interleaving, and recoverability, other gates decide whether to retrieve from uncertainty estimates \citep{moskvoretskii-etal-2025-adaptive}, from probes of intermediate hidden states \citep{baek-etal-2025-probing}, or from features that do not query the generator at all \citep{marina-etal-2025-llm}, and RetrievalQA benchmarks such gates on questions that require external knowledge \citep{zhang-etal-2024-retrievalqa}. Related work resolves conflicting evidence \citep{nie2026evotrustrag} or detects unanswerable questions from model internals \citep{lavi2025unanswerability}.

\paragraph{Selective prediction and abstention.} Extensions of selective classification reject out-of-distribution inputs \citep{narasimhan2024plugin}, enforce group fairness while abstaining \citep{yin2024fair}, derive optimal selectors from the Neyman–Pearson lemma under covariate shift \citep{heng2026abstain}, and calibrate sampling and filtering to control a specified risk of LLM answers \citep{wang2026safer}. Conformalized abstention policies adapt distribution-free risk control to language models \citep{tayebati2026cap}. For LLMs specifically, abstention has been surveyed \citep{wen-etal-2025-know}, benchmarked across question types and domains \citep{madhusudhan-etal-2025-llms}, and framed as a decision problem for healthcare \citep{presacan2026silence}.

\section{Proofs}
\label{app:proofs}

\paragraph{Proof of Proposition~\ref{prop:size}.}
Fix a base query $b$ with distractor set $D_b$. By construction, every state of $b$ retains exactly $|D_b|$ units, so a size-only estimator assigns one common score $f_b$ to all states of $b$, and the compression ratio $|D_b|/|H_b|$ is likewise constant within $b$. Let $n_u(b)$ and $n_c(b)$ be the numbers of unsafe and complete examples of $b$, and $P_u(b) = n_u(b)/\sum_{b'} n_u(b')$, $P_c(b) = n_c(b)/\sum_{b'} n_c(b')$. Unsafe AUROC is the probability that a uniformly drawn unsafe example outscores a uniformly drawn complete example, counting ties as one half. With $K(b,b') = \mathbb{1}\{f_b > f_{b'}\} + \tfrac12 \mathbb{1}\{f_b = f_{b'}\}$, this gives
\[
\mathrm{AUROC}(f) = \sum_{b,b'} P_u(b)\, P_c(b')\, K(b,b').
\]
If $P_u = P_c$, the two draws are exchangeable and $K(b,b') + K(b',b) = 1$, so AUROC $= 1/2$, including ties. In general, let $g(b) = \sum_{b'} P_c(b') K(b,b') \in [0,1]$. Since $\sum_b P_c(b) g(b) = 1/2$ by the same symmetry,
\[
\left|\mathrm{AUROC}(f) - \tfrac12\right| = \left|\sum_b \bigl(P_u(b) - P_c(b)\bigr) g(b)\right| \le \mathrm{TV}(P_u, P_c).
\]
Equal positive multiplicities $(n_u, n_c)$ for every base query make both distributions uniform, so AUROC $= 1/2$ exactly. \hfill$\square$

\paragraph{The fraction of exceptional queries does not bound the deviation.} Suppose $99$ base queries each contribute one complete and one unsafe example, and one query contributes one complete and $1{,}000$ unsafe examples. A size-only score that is larger on the exceptional query has AUROC $\tfrac12 + \tfrac12\bigl(\tfrac{1000}{1099} - \tfrac{1}{100}\bigr) > 0.94$, although only $1\%$ of queries are exceptional. The bound must therefore be stated in terms of $P_u$ and $P_c$.

\paragraph{MuSiQue.} The processed MuSiQue data contain $21{,}902$ complete, $21{,}902$ missing, and $21{,}493$ relation-lost variants, so every base query contributes one complete example and either one or two unsafe examples, and a fraction $\epsilon = 409/21{,}902$ contributes one. Then $P_c$ is uniform and
\[
\mathrm{TV}(P_u, P_c) = \frac{\epsilon(1-\epsilon)}{2-\epsilon} \approx 0.0092.
\]
With only two groups of queries, $P_u$ and $P_c$ agree up to scale within each group, so only comparisons between the groups can deviate from one half, and the deviation is at most half this total variation, $0.0046$. This dataset-wide value is a reference for the test split, whose exact fraction of one-unsafe queries depends on the hash assignment; the observed test deviation is $0.0007$.

\begin{proposition}[Coarse gates are suboptimal]\label{prop:coarse}
Fix a coverage $c$ and let $Y$ indicate that the reader answers incorrectly. Among all gates that accept a fraction $c$ of examples, selective risk is minimized by accepting the examples with the smallest $\eta(q,M) = \Pr(Y = 1 \mid q, M)$. A gate that ranks by a binary variable $Z$, such as the ground-truth integrity state, attains this minimum for every $c$ only if $Y$ is conditionally independent of $(q,M)$ given $Z$; otherwise its risk is strictly larger for some $c$.
\end{proposition}

\paragraph{Proof of Proposition~\ref{prop:coarse}.}
Let $\eta(q,M) = \Pr(Y = 1 \mid q, M)$ and let a gate accept a set $A$ with $\Pr(A) = c$. Selective risk is $\mathbb{E}[\eta \mid A]$, so minimizing it over all sets of measure $c$ accepts the $c$-fraction with the smallest $\eta$, by the standard exchange argument: if $x \in A$, $y \notin A$, and $\eta(x) > \eta(y)$, swapping them weakly decreases $\mathbb{E}[\eta \mid A]$. A gate that ranks by a binary $Z$ can only accept unions of $\{Z = 0\}$ and arbitrary subsets of $\{Z = 1\}$ (or conversely), so within a level set of $Z$ it selects without regard to $\eta$. Its risk equals the optimum for every $c$ only if $\eta$ is almost surely constant on each level set, that is, $Y \perp (q,M) \mid Z$; otherwise there exists $c$ for which some accepted $x$ and rejected $y$ share a level set with $\eta(x) > \eta(y)$, and the inequality is strict. \hfill$\square$

\section{Experimental Details}
\label{app:experimental}

\subsection{Datasets and Exclusions}
\label{app:data}

\paragraph{HotpotQA.} HotpotQA \citep{yang2018hotpotqa} builds about 113K question-answer pairs from Wikipedia articles and provides sentence-level supporting-fact annotations that mark which sentences contain the information needed to answer each question. We use the distractor-setting training and development records ($97{,}852$ in total), since the fullwiki validation context omits gold paragraphs for $5{,}316$ of $7{,}405$ questions; the test split has no public labels.

\paragraph{2WikiMultiHopQA.} 2WikiMultiHopQA \citep{ho2020constructing} contains $192{,}606$ questions built from Wikipedia and Wikidata. Wikipedia provides the text a model reads, and Wikidata provides structured information about entities and their relations, so each example carries an explicit reasoning path as evidence triples. Its templated construction makes the evidence entities easy to identify from the question, which is consistent with the saturation we observe (Section~\ref{sec:transfer}). Its test split ships without supporting facts, so we exclude it.

\paragraph{MuSiQue.} MuSiQue \citep{trivedi2022musique} contains about 25K answerable multi-hop questions in its main version. It starts from single-hop questions drawn from existing QA datasets and composes them into reasoning chains of two to four steps, where the output of one step is needed for the next. MuSiQue-Full adds unanswerable questions whose contexts closely match their answerable counterparts but lack the paragraph for one reasoning step. We exclude these unanswerable records from the main data, because they share a question identifier with their answerable counterparts, and reuse them as family B.

\paragraph{Exclusions.} Table~\ref{tab:data} summarizes the processed data. MuSiQue excludes $22{,}355$ unanswerable source records and $453$ questions with too few distractors; HotpotQA excludes $446$ questions with too few distractors; 2WikiMultiHopQA excludes $12{,}576$ test questions without supporting facts. Family B excludes $275$ pairs whose unanswerable context retains every supporting paragraph and $65$ whose contexts differ in length. Distractor removal and relation-loss title selection use salted SHA-256 orderings of the base question, so the selection is reproducible without stored randomness.

\begin{table}[t]
\caption{Processed evidence-QA datasets. Every raw record is used or excluded for a recorded reason, and every run uses the complete processed dataset.}
\label{tab:data}
\small
\begin{center}
\begin{tabular}{lrrrr}
\toprule
Dataset & Raw records & Base questions & Excluded & Variants \\
\midrule
MuSiQue & 44{,}710 & 21{,}902 & 22{,}808 & 65{,}297 \\
HotpotQA & 97{,}852 & 97{,}406 & 446 & 292{,}218 \\
2WikiMultiHopQA & 192{,}606 & 180{,}030 & 12{,}576 & 540{,}090 \\
\bottomrule
\end{tabular}
\end{center}
\end{table}

\subsection{Training and Evaluation}
\label{app:training}

The Stage B configuration (width 128, dropout 0.1, learning rate $3\times10^{-4}$) is selected on MuSiQue validation over a pre-specified eight-point grid by unsafe AUROC with a macro-F1 tie-break; all eight configurations fall between $0.963$ and $0.968$ validation AUROC. We report Wilcoxon signed-rank $p$-values with Holm correction descriptively and treat the bootstrap intervals as the inferential quantity. All runs use one NVIDIA RTX 3090, and every run records its runtime and memory provenance in the released artifacts.

\subsection{Pre-registered Criteria and Outcomes}
\label{app:hypotheses}

\paragraph{Hypotheses.} We fix five hypotheses before the final runs (Section~\ref{sec:setup}). \textbf{H1}: \MemSafe{} exceeds the matched mean-pooling comparator in unsafe AUROC on all three datasets with positive bootstrap intervals. \textbf{H2}: \MemSafe{} exceeds \LexSet{} and a flat TF-IDF logistic model on both primary metrics. \textbf{H3}: the unsafe-AUROC degradation from family A to family B is smaller for \MemSafe{} than for the flat TF-IDF and text-length baselines. \textbf{H4}: removing bridge-conditioned encoding reduces unsafe AUROC on MuSiQue and 2WikiMultiHopQA. \textbf{H5}: \MemSafe{} exceeds \LexSet{} in unsafe AUROC on MIMIC-IV~3.1.

\paragraph{Smoke gates.} Before any full run, all tests pass, all data invariants hold, the paragraph-count control sits at or below $0.60$ unsafe AUROC, the text-length control at or below $0.70$, and the proposed model exceeds the lexical estimator by at least $0.05$ on validation. H1 to H5 correspond to design-document hypotheses H6 to H10.

\paragraph{Outcomes.} H1 holds with a small effect; H2 holds; H3 is refuted; H4 holds on MuSiQue (Table~\ref{tab:ablation}) and is untested on 2WikiMultiHopQA, since the ablation runs on MuSiQue only; H5 is partially refuted, with \LexSet{} the better ranker at the smallest clinical scale and \MemSafe{} the better ranker at the two larger scales (Section~\ref{sec:clinical}).

\paragraph{Reader-gating criteria.} We pre-register the reader-gating study (Section~\ref{sec:reader}) separately, after H1 to H5 and before inspecting its results, with three criteria: (i) at a matched $10\%$ risk target, \MemSafe{} gating yields higher coverage than reader-confidence gating; (ii) \MemSafe{} gating yields lower unsafe accepted error than answering everything, at coverage matched to or higher than the memory-length control; (iii) the area under the risk-coverage curve (AURC) over answer correctness is lower for \MemSafe{} than for reader confidence. Criteria (i) and (ii) are not met as stated, and we do not count them as successes: answering every question is already wrong $85\%$ to $92\%$ of the time for both readers, none of the gates we evaluate reaches a reader-risk target between $5\%$ and $20\%$, and we replace them with fixed-coverage reporting after observing the base error rate. This does not show that the targets are unattainable for every gate: with $15\%$ of answers correct, an oracle that observes correctness could reach $5\%$ or $10\%$ coverage without error. Criterion (iii) holds: AURC is $0.7605$ (mean over three estimator seeds) for \MemSafe{}, against $0.8138$ for flat TF-IDF, $0.8184$ for reader confidence, $0.8306$ for \LexSet{}, and $0.8495$ for memory length. Apart from this replacement, which follows the observed base error rate rather than any gate's result, we do not change any criterion after observing its result.

\section{Additional Results}
\label{app:results}

\subsection{Stronger Baselines and External Transfer}
\label{app:hardening}

\begin{figure}[t]
\centering
\includegraphics[width=0.99\linewidth]{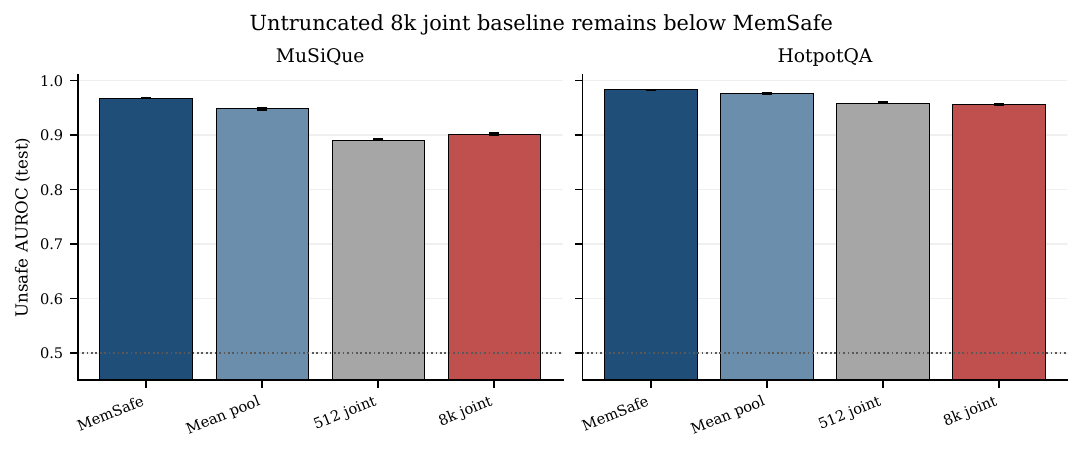}
\caption{The pinned 8k ModernBERT joint encoder improves on the original truncated concatenated baseline on MuSiQue but remains below \MemSafe{} and the matched mean-pooling comparator on both datasets. Bars show three-seed means with sample standard-deviation error bars.}
\label{fig:hardening-joint}
\end{figure}

\paragraph{8k joint encoder.} The pinned 149M-parameter ModernBERT reranker uses an $8{,}192$-token window on the identical question-memory serialization and objectives as the 512-token concatenated baseline, which truncates $100\%$, $98.8\%$, and $64.9\%$ of memories. Exact tokenization confirms that no MuSiQue or HotpotQA variant exceeds the window (maxima of $4{,}542$ and $3{,}464$ tokens). Across the three pre-specified seeds, it reaches mean unsafe AUROC and macro F1 of $0.9019$ and $0.5974$ on MuSiQue and $0.9559$ and $0.7534$ on HotpotQA (Figure~\ref{fig:hardening-joint}). At a validation-selected $5\%$ evidence risk, its mean coverage is $0.0028$ on MuSiQue and $0.1206$ on HotpotQA, against \MemSafe{}'s five-seed $0.2479$ and $0.2922$. An untruncated joint encoder therefore still underperforms; because the backbone also changes, this comparison does not isolate the effect of truncation on the concatenated baseline.

\paragraph{LLM judge.} We evaluate Qwen2.5-7B-Instruct, pinned to a public revision, by scoring $\Pr(\text{No})/(\Pr(\text{Yes})+\Pr(\text{No}))$ at the answer position on a deterministic $1{,}000$-question MuSiQue subset. The zero-shot pass uses an $8{,}192$-token context, truncates none of the $2{,}973$ test prompts, and reaches $0.7832$ unsafe AUROC ($0.8557$ AUPRC). The deterministic four-shot version draws its exemplars from the training split, uses $16{,}384$ tokens, likewise truncates none, and reaches $0.7825$ AUROC ($0.8686$ AUPRC), a matched change of $-0.0008$. On the same subset, \MemSafe{} reaches $0.9719$, the mean-pooling comparator $0.9537$, and flat TF-IDF $0.6975$. \MemSafe{}, with about 22M parameters (roughly $1/300$ of the judge), therefore exceeds the judge by $0.1887$ AUROC. The judge in turn outperforms flat TF-IDF, so a TF-IDF approximation underestimates autorater-style baselines. This comparison concerns the tested prompts and checkpoint, not LLM judges in general.
\label{app:judge}

\begin{figure}[t]
\centering
\includegraphics[width=0.72\linewidth]{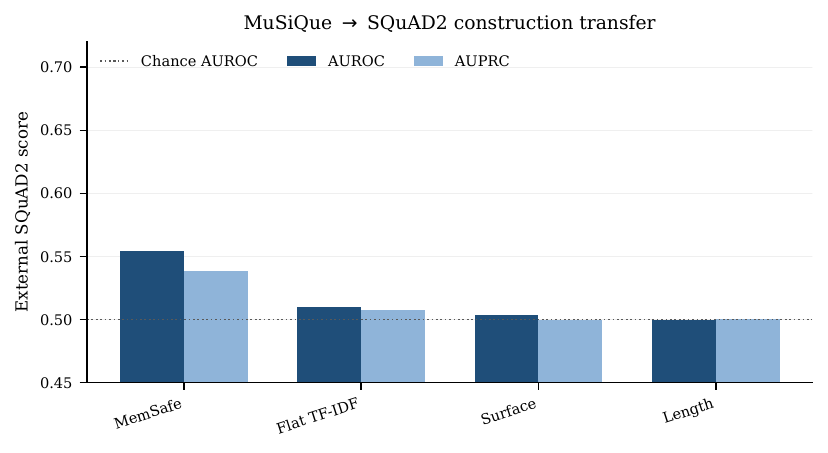}
\caption{Independent-construction transfer from MuSiQue to SQuAD~2.0. \MemSafe{} retains positive discrimination and exceeds the controls, but the absolute performance gap exposes the generalization limitation.}
\label{fig:hardening-external}
\end{figure}

\paragraph{SQuAD~2.0 transfer.} The $11{,}873$ development examples contain $5{,}928$ answerable and $5{,}945$ author-marked unanswerable questions. The frozen \MemSafe{} unsafe score reaches $0.6393$ AUROC, $0.6240$ AUPRC, $0.2915$ Brier score, and $0.0669$ ECE, against $0.5099$ AUROC for flat TF-IDF and exactly $0.5000$ for the memory-length control (Figure~\ref{fig:hardening-external}). Thresholds selected on MuSiQue validation do not transfer: at $5\%$ and $10\%$ evidence-risk budgets, they accept only $0.0049$ and $0.0479$ of examples, yet realize risks of $0.2759$ and $0.2900$, far above the targets. The 256-token pair encoder of \MemSafe{}, the same cap as in the main QA runs, truncates $4.79\%$ of the question-context pairs. The text-length control reaches $0.526$ train and $0.515$ development unsafe AUROC on SQuAD~2.0 (\emph{unsafe} meaning unanswerable), close to chance and consistent with our own text-length rows.

\subsection{Component Ablations}
\label{app:ablation}

\begin{figure}[t]
\centering
\includegraphics[width=0.99\linewidth]{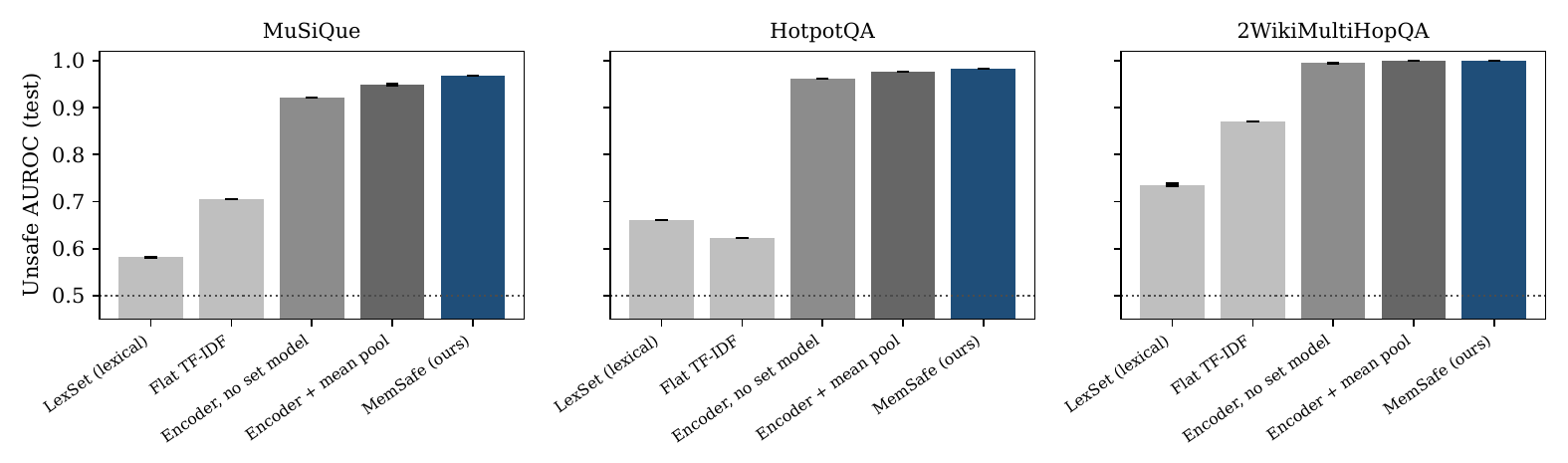}
\caption{Unsafe AUROC on the test split for the attribution ladder (five-seed means with seed-level standard deviations; the dotted line marks chance). The first jump combines architecture and fine-tuning; Figure~\ref{fig:interaction} and Table~\ref{tab:ablation} separate them on MuSiQue.}
\label{fig:attribution}
\end{figure}

Figure~\ref{fig:attribution} orders the attribution ladder from \LexSet{} to the full model; its rungs differ in several ways at once. Table~\ref{tab:ablation} (Section~\ref{sec:component-ablations}) instead removes one component at a time from the full estimator, holding the encoder, data, and heads fixed and varying only the Stage B seed (three seeds, one shared encoder), so each row isolates its own component.

\begin{figure}[t]
\centering
\includegraphics[width=0.9\linewidth]{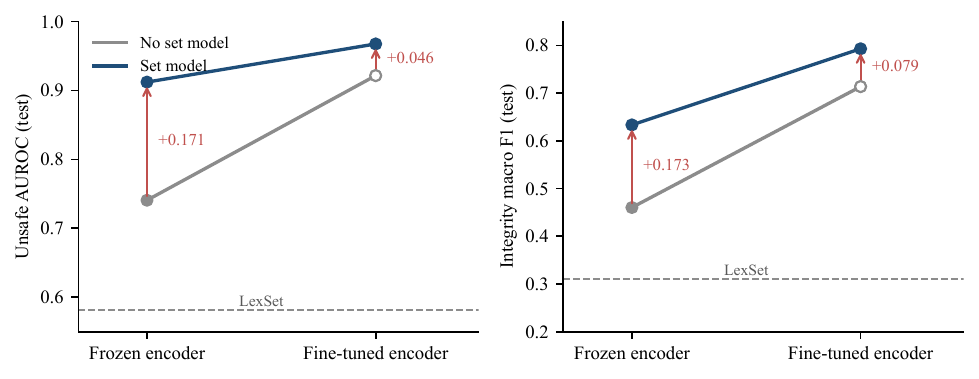}
\caption{Fine-tuning and the set model provide overlapping gains (MuSiQue, test split). Arrows show the gain from adding the set model: $+0.171$ unsafe AUROC with a frozen encoder and $+0.046$ with a fine-tuned one. Filled points average three Stage B seeds that share one encoder (Table~\ref{tab:ablation}); the hollow point is the five-seed comparator from Table~\ref{tab:main}, so the figure mixes two replication scopes. The dashed line marks \LexSet{}.}
\label{fig:interaction}
\end{figure}

The last row of Table~\ref{tab:ablation} removes both fine-tuning and the set model, so a frozen pretrained cross-encoder supplies aggregated relevance scores to a logistic head. This model reaches $0.7407$, against $0.5820$ for \LexSet{} and $0.9674$ for the full model. A frozen pretrained cross-encoder therefore already closes $0.1587$ of the $0.3854$ gap ($41\%$). Because later rungs change several components at once, this ratio describes the ladder rather than allocating credit causally among components.

Comparing this row with the frozen-plus-set-model row and with the fine-tuned, no-set-model comparator of Table~\ref{tab:main} ($0.9215$) shows that fine-tuning and the set model are not additive. Fine-tuning's marginal contribution is $0.9215 - 0.7407 = 0.1808$ when added to the no-set-model baseline, but only $0.9674 - 0.9121 = 0.0553$ when added on top of the set model. The set model's marginal contribution is symmetric: $0.9121 - 0.7407 = 0.1714$ when added to the frozen, no-fine-tuning baseline, but only $0.9674 - 0.9215 = 0.0459$ when added on top of fine-tuning. Each component contributes three to four times more when the other is absent, which suggests that both capture largely the same evidence signal. Because the fine-tuned, no-set-model value ($0.9215$) comes from the five-seed comparator rather than the three-seed shared-encoder panel, this is a descriptive ladder, not a matched factorial interaction estimate. Bridge-conditioned encoding is the second-largest single-component ablation ($-0.0251$), consistent with its purpose of exposing second-hop evidence that the question does not mention lexically; this supports H4 on MuSiQue, while the ablation does not run on 2WikiMultiHopQA, where H4 is also registered (Appendix~\ref{app:hypotheses}). The lexical and position features inherited from \LexSet{} contribute little once cross-encoded representations are available ($-0.0038$).

Removing the per-unit evidence and missing-count heads changes neither metric beyond seed noise ($+0.0002$ and $-0.0014$). These heads provide localization outputs (per-unit AUPRC of $0.9123$) rather than improving detection.

\subsection{Silver Supervision}
\label{app:silver}

The silver per-unit target marks a training unit as evidence when the gold answer string appears in its text, a standard distant-supervision heuristic. Only the per-unit targets of the training split change; validation, test, and integrity labels are unchanged. Compared with gold labels, the silver target has $0.653$ precision and $0.365$ recall on the training split. Over three seeds, the silver-supervised model reaches $0.9444 \pm 0.0021$ unsafe AUROC on test, three points below the gold-supervised $0.9674 \pm 0.0008$ and still $0.36$ above \LexSet{} ($0.582$); macro F1 falls from $0.7929$ to $0.7175$, and per-unit evidence AUPRC falls from $0.9123$ to $0.4252$. Gold unit targets are thus largely unnecessary for detection but remain important for per-unit localization. Gold supporting facts still define the memory variants and integrity labels in both conditions, so the experiment concerns noisy unit targets within a gold-constructed benchmark.

\subsection{Construction Transfer and Memorization}
\label{app:memorization}

\begin{figure}[t]
\centering
\includegraphics[width=0.99\linewidth]{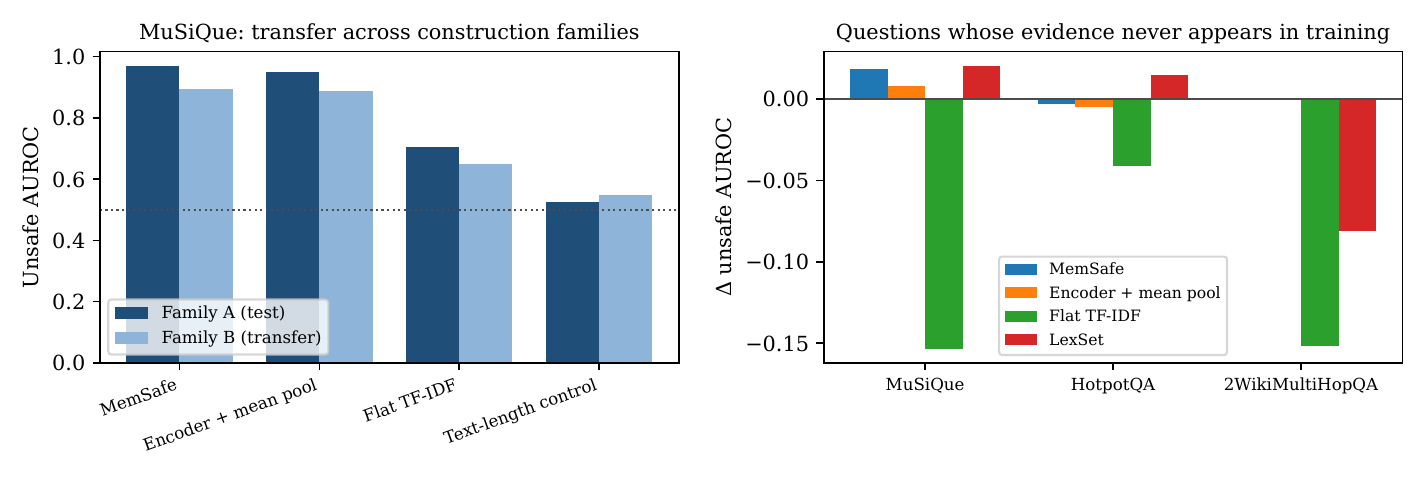}
\caption{Left: unsafe AUROC on the size-matched family A and on MuSiQue-Full's native contrasts (family B). Right: change in unsafe AUROC when evaluation is restricted to questions whose gold evidence never appears in training. The lexical baselines lose accuracy; the proposed estimator does not.}
\label{fig:transfer}
\end{figure}

Figure~\ref{fig:transfer} shows the family-A-to-B transfer (left) and the memorization audit (right). Splits are disjoint by base question, but paragraphs recur across questions: $77.0\%$ (MuSiQue), $54.4\%$ (HotpotQA), and $62.5\%$ (2WikiMultiHopQA) of test paragraphs also appear in training, and among shared paragraphs the evidence status is identical in $76\%$ to $87\%$ of cases. A model could therefore memorize paragraph-level evidence status instead of judging sufficiency for the query. Restricting evaluation to questions whose gold evidence never appears in training changes \MemSafe{} by $+0.019$, $-0.003$, and $-0.000$, while flat TF-IDF falls by up to $0.153$ and \LexSet{} by $0.081$ on 2WikiMultiHopQA. Shared evidence paragraphs therefore do not explain the performance of \MemSafe{}, whereas the lexical baselines benefit from them. The restriction also changes the evaluated question population and leaves pretraining overlap untested, so it does not exclude every form of memorization.

\subsection{Selective Prediction and Errors}
\label{app:selective}

\begin{figure}[t]
\centering
\includegraphics[width=0.99\linewidth]{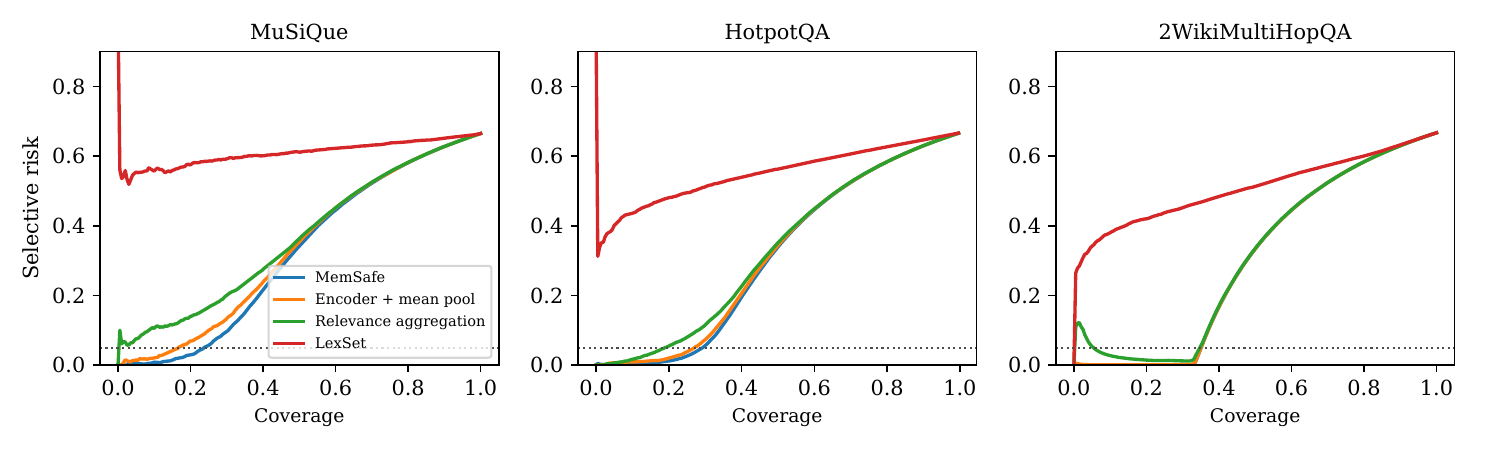}
\caption{Risk-coverage curves on the test split (seed 17). The dotted line marks a $5\%$ evidence-aware risk budget. Curves that reach further right before crossing it support higher usable coverage.}
\label{fig:riskcoverage}
\end{figure}

Table~\ref{tab:selective} and Figure~\ref{fig:riskcoverage} report the full selective-prediction results for Section~\ref{sec:selective}. On MIMIC-IV outpatient measurement records (OMR), neither estimator reaches useful coverage at the $5\%$ budget ($3.1\%$ for \LexSet{} and $0.2\%$ for \MemSafe{}).

\paragraph{Prevalence limits on evidence risk.} Let $\pi$ be the unsafe prevalence. A gate that ranks every complete example before every unsafe one has evidence risk $R_E^*(c) = \max\{0,\, 1 - (1-\pi)/c\}$ at coverage $c$. No gate can accept more than $1-\pi$ complete mass, so a target $R_E \le \alpha < 1$ implies
\[
c \le \min\left\{1,\, \frac{1-\pi}{1-\alpha}\right\},
\qquad
\mathrm{AURC}_{\min} = \pi + (1-\pi)\log(1-\pi),
\]
where the second quantity integrates $R_E^*$ over coverage. HotpotQA and 2WikiMultiHopQA contribute exactly three variants per question, so $\pi = 2/3$, the coverage ceiling at $\alpha = 5\%$ is $0.3509$, and $\mathrm{AURC}_{\min} \approx 0.30046$. On 2WikiMultiHopQA, \MemSafe{} reaches $0.3509$ coverage and $0.3005$ AURC, and mean pooling $0.3502$ and $0.3008$: both are at the prevalence limit, so their differences there do not reflect ranking quality. The same limit explains why near-perfect AUROC coexists with AURC near $0.30$, and AURC values should not be compared across populations with different prevalence.

\begin{table}[t]
\caption{Selective prediction with an evidence-aware risk definition. AURC is lower-is-better; coverage is the test fraction accepted at a threshold chosen on validation for a $5\%$ risk target, and the last column is the risk realized at that threshold. n/a: no realized risk is reported for this row. Bold marks the best value within each dataset; $\downarrow$ and $\uparrow$ mean lower and higher is better. Realized risk should be close to the target rather than minimal, so it is not bolded. 2WikiMultiHopQA values are not bolded because both encoder models sit at the prevalence limits (coverage $0.3509$, AURC $0.30046$; Appendix~\ref{app:selective}).}
\label{tab:selective}
\small
\begin{center}
\begin{tabular}{llccc}
\toprule
Dataset & Model & AURC $\downarrow$ & Coverage at $5\%$ target $\uparrow$ & Realized test risk \\
\midrule
MuSiQue & \MemSafe{} & $\mathbf{0.3197}$ & $\mathbf{0.2479}$ & $0.0553$ \\
 & Encoder + mean pool & $0.3359$ & $0.1334$ & $0.0463$ \\
 & Relevance aggregation & $0.3626$ & $0.0206$ & $0.0627$ \\
 & \LexSet{} & $0.6088$ & $0.0002$ & $0.7333$ \\
\midrule
HotpotQA & \MemSafe{} & $\mathbf{0.3120}$ & $\mathbf{0.2922}$ & $0.0505$ \\
 & Encoder + mean pool & $0.3166$ & $0.2750$ & $0.0514$ \\
 & \LexSet{} & $0.5495$ & $0.0000$ & n/a \\
\midrule
2WikiMultiHopQA & \MemSafe{} & $0.3005$ & $0.3509$ & $0.0502$ \\
 & Encoder + mean pool & $0.3008$ & $0.3502$ & $0.0487$ \\
 & \LexSet{} & $0.5075$ & $0.0000$ & n/a \\
\midrule
MIMIC-IV OMR & \LexSet{} & $\mathbf{0.4416}$ & $\mathbf{0.0311}$ & n/a \\
 & \MemSafe{} & $0.4924$ & $0.0017$ & n/a \\
\bottomrule
\end{tabular}
\end{center}
\end{table}

\paragraph{Error taxonomy.}
\label{app:errors}
\MemSafe{}'s test error rate is $0.1985$, $0.1396$, and $0.0103$ on MuSiQue, HotpotQA, and 2WikiMultiHopQA. Errors concentrate among unsafe states rather than across the safety boundary: relation-lost predicted as missing accounts for $28.0\%$, $33.6\%$, and $42.3\%$ of errors, and missing predicted as relation-lost for a further $26.6\%$, $20.6\%$, and $23.5\%$. Unsafe memories predicted complete, the safety-relevant direction, make up $6.6\%$, $6.0\%$, and $0.4\%$ of unsafe cases, while complete memories are rejected at $13.8\%$, $7.2\%$, and $0.3\%$, so the estimator errs toward caution. Difficulty tracks how much evidence survives rather than how many hops a question needs: the error rate peaks when exactly one required item remains ($0.280$, $0.245$, and $0.026$ on the three datasets), against $0.151$ to $0.174$ for zero or two surviving items on MuSiQue. Most remaining errors therefore lie at the boundary between relation-lost and missing memory, a distinction that also matters for targeted retrieval.

\subsection{Reader Gating}
\label{app:reader}

Table~\ref{tab:reader} reports the gating results for the 7B reader, which Figure~\ref{fig:clinicalreader} (right) plots.

\paragraph{LLM judge as a gate.} With the 7B reader, the zero-shot judge's error among answered questions is $0.6419$, $0.6700$, $0.7389$, and $0.7786$ at $5\%$, $10\%$, $25\%$, and $50\%$ coverage. \MemSafe{} is better at $25\%$ and $50\%$ coverage on every seed, the judge is better at $10\%$ on every seed, and at $5\%$ coverage the mean favors \MemSafe{} narrowly ($0.6306$ against $0.6419$) while one of three seeds ($0.6486$) does not. This pattern is consistent with Proposition~\ref{prop:coarse}: a gate should rank by $\eta(q,M)$ rather than by integrity, and a judge from the reader's model family may carry information about which questions that reader can answer, which an integrity estimator does not observe. The two signals may therefore be complementary; we do not isolate a model-family effect or test a combined gate.

\begin{figure}[t]
\centering
\includegraphics[width=0.99\linewidth]{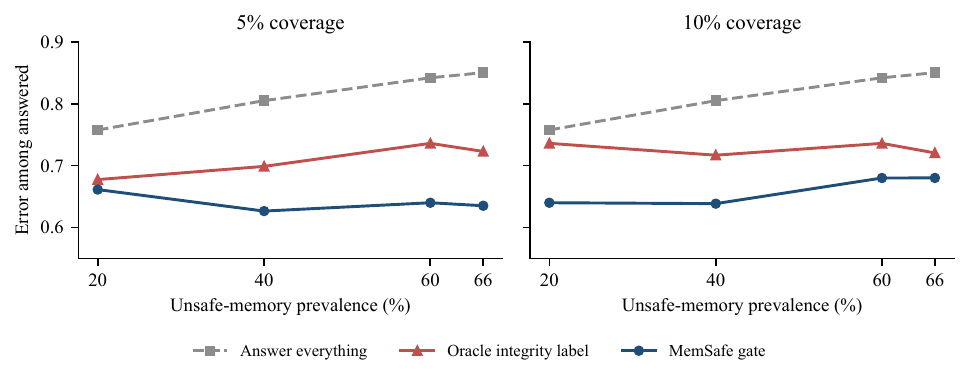}
\caption{Reader gating under lower unsafe-memory prevalence (MuSiQue, 7B reader, one estimator seed). Complete memories stay fixed and unsafe memories are subsampled; $66\%$ is the construction's native rate. Error among answered questions at $5\%$ (left) and $10\%$ (right) coverage: the \MemSafe{} gate lies below both the oracle integrity label and answering everything at every prevalence tested.}
\label{fig:prevalence}
\end{figure}

\paragraph{Lower corruption prevalence.} The headroom in Section~\ref{sec:reader} depends on the construction's native ratio of one complete to two unsafe memories ($66.4\%$ unsafe), which need not match the rate in deployment. We hold every complete-memory example fixed and subsample unsafe examples to prevalences of $20\%$, $40\%$, and $60\%$ (one estimator seed; Figure~\ref{fig:prevalence}). At $5\%$ coverage, gating on the predicted score beats both answering everything and the oracle label at every prevalence, from $20\%$ ($0.661$ gated, $0.758$ answer-all, $0.677$ oracle) to the native $66\%$ ($0.635$, $0.850$, and $0.723$); the same ordering holds at $10\%$ coverage. The absolute gain over answering everything shrinks as prevalence falls, because fewer memories are corrupted, but the ordering does not change.

\begin{table}[t]
\caption{Selective answering on MuSiQue with the 7B reader: error among answered questions at fixed coverage (lower is better). Answering everything is wrong $85.03\%$ of the time. \MemSafe{} and \LexSet{} use three independently trained estimator seeds (mean $\pm$ sample standard deviation); every other row is deterministic given the reader and the subset, so it has no seed variance. The oracle row gates on the true integrity label, which is not an oracle for answer correctness. The LLM judge is the better gate at $10\%$ coverage. Bold marks the lowest error in each column.}
\label{tab:reader}
\small
\begin{center}
\begin{tabular}{lcccc}
\toprule
& \multicolumn{4}{c}{Error among answered questions $\downarrow$} \\
\cmidrule(lr){2-5}
Gate & $5\%$ coverage & $10\%$ & $25\%$ & $50\%$ \\
\midrule
\MemSafe{} unsafe score & $\mathbf{0.6306 \pm 0.0206}$ & $0.6824 \pm 0.0039$ & $\mathbf{0.7084 \pm 0.0069}$ & $\mathbf{0.7687 \pm 0.0027}$ \\
LLM judge (zero-shot, 7B) & $0.6419$ & $\mathbf{0.6700}$ & $0.7389$ & $0.7786$ \\
Oracle integrity label & $0.7230$ & $0.7205$ & $0.7133$ & $0.7793$ \\
Reader sequence confidence & $0.7027$ & $0.7744$ & $0.8398$ & $0.8264$ \\
Flat TF-IDF logistic & $0.7838$ & $0.7845$ & $0.8008$ & $0.8176$ \\
\LexSet{} unsafe score & $0.7995 \pm 0.0103$ & $0.8047 \pm 0.0067$ & $0.8259 \pm 0.0047$ & $0.8452 \pm 0.0013$ \\
Memory-length control & $0.8649$ & $0.8788$ & $0.8533$ & $0.8392$ \\
\bottomrule
\end{tabular}
\end{center}
\end{table}

\subsection{Clinical Results}
\label{app:clinical}

\paragraph{Setup.} Longitudinal clinical memories are numeric measurement events whose integrity depends on temporal order and supersession. Each MIMIC-IV~3.1 source runs the full model suite with five seeds, deterministic patient sampling, and a locked data selection that every seed must match. The smallest setting reproduces the configuration of the prior lexical system. Table~\ref{tab:clinical} reports the three clinical settings, and Table~\ref{tab:clinical_curve} and Figure~\ref{fig:clinicalreader} (left, Section~\ref{sec:clinical}) report the learning curve that locates the crossover.

\paragraph{Largest scale.} On laboratory events, \MemSafe{} reaches $0.9908 \pm 0.0002$ unsafe AUROC and $0.9575 \pm 0.0008$ macro F1, against $0.9567 \pm 0.0091$ and $0.9095 \pm 0.0031$ for \LexSet{} and $0.8852$ and $0.8538$ for the rule baseline; on scale-matched OMR it reaches $0.9960 \pm 0.0003$ and $0.9726 \pm 0.0013$. \LexSet{} encodes numeric trends, event positions, and inter-event gaps as explicit features and therefore needs little data to rank unsafe memory, which makes it preferable for small cohorts.

\begin{table}[t]
\caption{MIMIC-IV test-split results over five seeds. The first two rows share an event source but differ in patient sampling fraction and question cap, so the change in sign points to a scale effect rather than a modality effect without isolating training-data quantity from cohort composition. Positive differences favor \MemSafe{}. Bold marks the better of the two estimators in each row; $\uparrow$ means higher is better.}
\label{tab:clinical}
\small
\begin{center}
\resizebox{\linewidth}{!}{%
\begin{tabular}{lrrcccc}
\toprule
Setting & Patients & Train variants & \MemSafe{} AUROC $\uparrow$ & \LexSet{} AUROC $\uparrow$ & $\Delta$ AUROC (seed wins) & $\Delta$ Macro F1 \\
\midrule
OMR, $1/64$ patients, 1k cap & 683 & 2{,}780 & $0.9220$ & $\mathbf{0.9649}$ & $-0.0429$ (0/5) & $+0.0880$ \\
OMR, $1/16$ patients, 60k cap & 3{,}934 & 29{,}920 & $\mathbf{0.9960}$ & $0.9725$ & $+0.0236$ (5/5) & $+0.1475$ \\
Laboratory events, $1/16$, 60k cap & 7{,}869 & 166{,}560 & $\mathbf{0.9908}$ & $0.9567$ & $+0.0341$ (5/5) & $+0.0480$ \\
\bottomrule
\end{tabular}
}
\end{center}
\end{table}

\begin{table}[t]
\caption{Locating the sample-efficiency crossover: a five-point learning curve on MIMIC-IV OMR test-split results, holding the event source and patient sampling fraction fixed at $1/16$ and varying only the base-question cap (three seeds per point, fewer than the five seeds used elsewhere); realized patient counts grow with the cap. At $1/16$ sampling, the $20{,}000$ and $60{,}000$ caps both exceed the available questions and yield the identical data selection (same selection hash), so the $20{,}000$-cap row reproduces the scale-matched OMR setting of Table~\ref{tab:clinical} on a seed subset ($0.9957$ versus $0.9960$ unsafe AUROC for \MemSafe{}; \LexSet{} matches at $0.9725$ in both). Positive differences favor \MemSafe{}. Bold marks the better of the two estimators in each row; $\uparrow$ means higher is better.}
\label{tab:clinical_curve}
\small
\begin{center}
\resizebox{\linewidth}{!}{%
\begin{tabular}{rrrcccc}
\toprule
Cap & Patients & Train variants & \MemSafe{} AUROC $\uparrow$ & \LexSet{} AUROC $\uparrow$ & $\Delta$ AUROC (seed wins) & $\Delta$ Macro F1 \\
\midrule
1{,}000 & 929 & 2{,}788 & $0.9462$ & $\mathbf{0.9626}$ & $-0.0163$ (1/3) & $+0.0991$ \\
2{,}500 & 2{,}002 & 6{,}844 & $\mathbf{0.9911}$ & $0.9648$ & $+0.0263$ (3/3) & $+0.1774$ \\
5{,}000 & 3{,}113 & 13{,}724 & $\mathbf{0.9935}$ & $0.9704$ & $+0.0231$ (3/3) & $+0.1733$ \\
10{,}000 & 3{,}906 & 27{,}788 & $\mathbf{0.9955}$ & $0.9736$ & $+0.0219$ (3/3) & $+0.1614$ \\
20{,}000 & 3{,}934 & 29{,}920 & $\mathbf{0.9957}$ & $0.9725$ & $+0.0232$ (3/3) & $+0.1181$ \\
\bottomrule
\end{tabular}
}
\end{center}
\end{table}

The finer sweep confirms the ordering of Table~\ref{tab:clinical} at both ends and narrows the crossover from a two-point bracket ($2{,}780$ to $29{,}920$ training variants) to the interval between $2{,}788$ and $6{,}844$: \LexSet{} still wins two of three seeds at the former and \MemSafe{} wins all three at the latter, with a stable AUROC gap from $6{,}844$ variants onward and macro F1 favoring \MemSafe{} at every point on the curve.

\subsection{Additional Diagnostics}
\label{app:diagnostics}

\paragraph{Macro-F1 differences.} Against the mean-pooling comparator, \MemSafe{} improves integrity macro F1 by $+0.0295$, $+0.0176$, and $+0.0014$ on MuSiQue, HotpotQA, and 2WikiMultiHopQA.

\paragraph{Localization and calibration.} \MemSafe{}'s per-unit evidence AUPRC is $0.9123$, $0.9668$, and $0.9998$ on MuSiQue, HotpotQA, and 2WikiMultiHopQA, against $0.8850$, $0.9474$, and $0.9994$ for the mean-pooling comparator; its missing-count MAE is $0.3764$, $0.1957$, and $0.0457$, against $0.4634$, $0.2431$, and $0.0613$. Its ECE is $0.0128$, $0.0065$, and $0.0013$.

\paragraph{Seed variability.} Table~\ref{tab:seedvar} reports seed-level variability for the learned models. Deterministic baselines (majority, paragraph count, text length, flat TF-IDF, and retrieval similarity) have zero seed variance, because their training procedure contains no random state.

\begin{table}[t]
\caption{Seed-level standard deviation of test unsafe AUROC for learned models (five seeds; three for the concatenated cross-encoder). Lower means more stable across seeds ($\downarrow$); bold marks the lowest value in each column.}
\label{tab:seedvar}
\small
\begin{center}
\begin{tabular}{lccc}
\toprule
Model & MuSiQue $\downarrow$ & HotpotQA $\downarrow$ & 2WikiMultiHopQA $\downarrow$ \\
\midrule
\MemSafe{} & $\mathbf{0.0003}$ & $0.0003$ & $\mathbf{0.0000}$ \\
Encoder + mean pool & $0.0018$ & $0.0002$ & $\mathbf{0.0000}$ \\
Relevance aggregation & $0.0008$ & $\mathbf{0.0001}$ & $0.0005$ \\
Concatenated cross-encoder & $0.0009$ & $0.0009$ & $0.0001$ \\
\LexSet{} & $0.0009$ & $0.0002$ & $0.0031$ \\
\bottomrule
\end{tabular}
\end{center}
\end{table}

%

\end{document}